# Quantifying the human visual exposome with vision language models

Christian Rominger [(1)], Andreas R. Schwerdtfeger [(1)], Malay Gaherwar Singh [(2)], Dimitri Khudyakow [(2)], Elizabeth A. M. Michels [(2)], Fabian Wolf [(2)], Jakob Nikolas Kather [(2,3,4)*], Magdalena Katharina Wekenborg [(2)*]

*equal last authorship

(1) Health Psychology, University of Graz, Graz, Austria

(2) Else Kröner Fresenius Center for Digital Health, Faculty of Medicine and University Hospital Carl Gustav Carus, TUD Dresden University of Technology, Dresden, Germany

(3) Department of Medicine I, Faculty of Medicine and University Hospital Carl Gustav Carus, TUD Dresden University of Technology, Dresden, Germany

(4) Medical Oncology, National Center for Tumor Diseases (NCT), University Hospital Heidelberg, Heidelberg, Germany

**Corresponding authors**

Jakob Nikolas Kather, MD, MSc; Professor of Clinical Artificial Intelligence Else Kröner Fresenius Center for Digital Health TUD Dresden University of Technology, Fetscherstrasse 74 01307 Dresden, Germany; Email: kather.jn@tu-dresden.de

Magdalena Wekenborg, PhD, Research Group Leader Else Kröner Fresenius Center for Digital Health, TUD Dresden University of Technology, Fetscherstrasse 74 01307 Dresden, Germany; Email: Magdalena.wekenborg@tu-dresden.de

## Abstract

The visual environment is a fundamental yet unquantified determinant of mental health. While the concept of the environmental exposome is well-established, current methods rely on coarse geospatial proxies or biased self-reports, failing to capture the first-person visual context of daily life. We addressed this gap by coupling ecological momentary assessment with vision-language models (VLMs) to quantify the semantic richness of human visual experience. Across 2,674 participant-generated photographs, VLM-derived estimates of "greenness" robustly predicted momentary affect and chronic stress, consistent with established benchmarks. We then developed a semi-autonomous large language model (LLM)-based pipeline that mined over seven million scientific publications to extract nearly 1,000 environmental features empirically linked to mental health. When applied to real-world imagery, up to 33% of VLM-extracted context ratings significantly correlated with affect and stress. These findings establish a scalable, objective paradigm for visual exposomics, enabling high-throughput decoding of how the visible world shapes mental health.

## Main

Mental illness creates an immense global burden on public health and economies [1]. Decades of research suggest that everyday surroundings exert a profound influence on mental health [2], with greenness emerging as one of the most consistently replicated environmental predictors [3–6]. Collectively, these surroundings can be summarized as the "environmental exposome" [7], with the visual perception of the environment termed the visual exposome. Still, our ability to understand and quantify how the visual exposome shapes mental health remains surprisingly limited: we can measure where people are and how they subjectively perceive their surroundings, but we remain effectively blind to what people actually see in their daily lives. GPS-based metrics and satellite-derived indices provide useful macro-level proxies of environmental features [4,8–10], but they cannot cover the semantic richness, spatial nuance, or social signals embedded in real-world visual scenes. Most crucially, they overlook the immediate, first-person environments individuals actually encounter, whether a space is crowded or quiet, whether animals or other people are present, whether natural light, greenery, or stress-inducing cues appear in view [11,12]. Self-report measures reflect only subjective impressions and are prone to response biases. As a result, the visual exposome, arguably the most proximal form of contextual exposure, has remained empirically inaccessible at scale.

An optimal way to quantify the visual exposome would be to use photographs taken by participants themselves to capture health-relevant contextual features of their environments. This method provides a large number of data points that reflect the participant's surroundings from a first-person perspective, allowing researchers to see the world through the participants' eyes. To date, however, this approach has been used almost exclusively in qualitative research, for example, to explore subjective experiences, health contexts, or to illustrate environmental features [13–15], but not in quantitative frameworks examining associations between visually captured aspects of daily environments and psychological and mental health-relevant outcomes. This limited application is most likely due to the reliance on manual expert evaluation of visual content according to predefined categories, an approach that is time-consuming, subjective, and inherently not scalable, restricting its use in intensive longitudinal studies with large sample sizes. Recent advances in vision-language models (VLMs) now allow us to automatically extract semantically meaningful features from images, as illustrated by their near-human-level scene understanding across domains such as autonomous driving [16,17].

Therefore, we hypothesized that VLMs could serve as scalable, objective tools to decode the complex visual information encoded in photographs. Embedded within an ecological momentary assessment framework [18], this could effectively unlock a new dimension of human phenotyping: continuous, first-person mapping of the visual exposome in its entirety.

As a first step, we applied VLMs to well-established greenness-related features (i.e., *greenness, nature score*, *plant presence, natural light exposure, inside/outside*) [12,19,20], using them as a benchmark to assess whether VLM-derived estimates from 2,674 participant-generated photographs correspond to participants' self-ratings of greenness and capture the well-documented associations with mental health-relevant outcomes reported in prior work [2,8,20].

Next, we mapped the entirety of environmental features previously linked to affect and stress. We developed and applied a semi-autonomous AI-based literature mining system that analyzed more than seven million scientific publications to identify nearly one thousand contextual features with empirical support for associations with affect and stress. These features were then used as objectives for VLM-based ratings, allowing us to examine whether the resulting VLM-derived contextual features can reproduce the empirically well-documented associations of the human exposome.

## Results

### Participant characteristics and study design

We conducted an ecological momentary assessment study combining self-reports, participant-generated photographs, and an AI-driven literature-mining approach to test whether VLMs can quantify a wide range of real-world environmental features relevant to mental health (Figure 1). Data from 106 adults (28 men; mean age = 24.9 ± 10.7 years) who participated in a 7-day ecological momentary assessment study using the *esmira* smartphone application were available [21]. Participants received seven randomly timed alarms per day between 09:00 a.m. and 08:00 p.m. to complete brief self-reports on indicators of mental health, namely positive and negative affect, and contextual information on greenness while simultaneously capturing a photograph of their current environment from their own perspective. In total, 2,674 participant-generated photographs in combination with self-reports were available for analysis. Participants further completed a questionnaire on sociodemographic variables and chronic stress. All subsequent analyses are based on this dataset. A detailed description of our methodology is provided in the Methods.

### VLMs quantify greenness and related visual features linked to mental health

As an initial step to validate the VLM framework against established evidence, we focused on greenness as a core visual property with well-established links to mental health. Beyond greenness, we quantified four visual environmental features that commonly accompany it, namely, *plant presence*, *nature score* (degree of natural scenery), *natural light exposure* (the intensity of daylight), and *inside/outside* classification, yielding five greenness-related visual measures derived from the VLMs. Each participant-generated photograph was subsequently rated 5 times on these dimensions by the open-weight LLaMA 4 VLM, which also provided a confidence score for every rating (see Supplementary Table 1 for more details). We then validated these model-derived ratings by examining their correspondence with participants' subjective experience of their surroundings. Multilevel analysis revealed significant associations between participants' self-reported greenness ratings and VLM-derived ratings of greenness itself and related visual features. More precisely, participants rated their surroundings as greener when their photographs were characterized by higher VLM-derived ratings of greenness ($t(2571.73) = 44.41$, $b = 1.26$, $p < 0.001$) and the other greenness-related features (i.e., *plant presence*, *natural light exposure*, *inside/outside;* see Table 1).

To assess whether these VLM-derived environmental features capture established relationships between greenness and mental health-related variables, we examined their associations with both momentary state-level (i.e., short-term experiences, namely positive and negative affect) and trait-level measures (more stable characteristics, namely mean affect over the study period and chronic stress). Across both time scales, VLM-derived visual greenness features were robustly associated with mental health-related variables (Table 2). Specifically, greater greenness was linked to higher momentary positive affect ($t(2564.8) = 2.64$, $b = 0.20$, $p = 0.008$). At the trait level, mental health-related variables also showed systematic relationships with greenness-related features (e.g., *natural light exposure* and positive affect: $t(105.6) = 2.31$, $b = 0.46$, $p = 0.022$; for more details see Table 2). Participants reporting higher average positive affect and lower negative affect over the study period tended to capture photographs containing more greenness, as reflected in higher *greenness* (positive affect: $t(104.7) = 2.57$, $b = 0.62$, $p = 0.012$, negative affect: $t(109.7) = -2.58$, $b = -1.07$, $p = 0.011$) and *nature score* (positive affect: $t(103.3) = 2.96$, $b = 0.63$, $p = 0.004$, negative affect: $t(108.2) = -2.25$, $b = -0.82$, $p = 0.026$) values. The average score of continuous VLM-derived greenness indicators was also significantly associated with higher trait positive affect ($t(104.6) = 2.75$, $b = 0.62$, $p = 0.007$). Being *inside/outside* (1 vs. 2) was linked with a higher positive affective state ($t(2557.0) =$ 3.26, $b = 0.04$, $p = 0.001$). Similarly, higher VLM-derived *greenness* ($r = -0.21$, $p = 0.031$) and *nature score* ($r = -0.23$, $p = 0.019$) were associated with lower perceived chronic stress at trait level. The other two VLM-derived greenness indicators as well as *inside/outside* showed negative effects of similar sizes (*rs* ≥ -0.18; see Figure 2) indicating comparable findings for all VLM-derived greenness indicators. The average score of continuous VLM-derived greenness indicators was significant with $r = -0.22$ ($p = 0.025$). Together, these data show that VLMs can reliably quantify health-relevant environmental features from real-world images and capture their meaningful associations with positive affect, negative affect, and stress.

**VLMs yield consistent contextual ratings linked to mental health across models**

To evaluate the generalizability of our findings across different model architectures, we compared the contextual feature ratings generated by LLaMA 4 VLM with those produced by the open-weight Qwen3 VL. As illustrated in Figure 2b, both models yielded highly similar ratings for *greenness*, *nature score*, and *plant presence* (*rs* = 0.83 - 0.89, all *ps* < 0.001). The correlation for *natural light exposure* ($r = 0.45$) and *inside/outside* was lower ($r = 0.66$; both with $p < 0.001$), suggesting slightly greater variability in this feature between models. As detailed in Supplementary Table 2, Qwen3 VL reproduced the overall pattern of associations between VLM-derived greenness-related features and mental health-relevant variables (i.e., state- and trait-level positive and negative affect) observed in the primary analyses with LLaMA 4 VLM (see Supplementary Table 2). Furthermore, *natural light exposure, plant presence*, and *greenness* showed a trend association with stress as a mental health-related variable, with similar effect sizes in the replication analyses (Supplementary Figure 1). These results confirm that VLM-derived contextual ratings are consistent across models to effectively capture mental health-relevant environmental features.

### The VLM framework generalizes to a broad literature-derived space of contextual features

To examine whether the framework extends beyond greenness-related features to a broader conceptual space of environmental determinants of mental health, we mapped environmental features previously linked to mental health using a large-scale, AI-driven literature mining approach. In a multiple-step process, we analyzed more than seven million open-access full-text articles from the Europe PMC database using the open-weight GPT-OSS-120B model, yielding a large set of 997 unique literature-derived contextual variables with empirical support for associations with affect and stress (prompts, code, and results are openly available; see Supplementary Figure 2). These identified features were then used as prompts for VLM-based ratings of the participant-generated photographs, following the same procedure as for the greenness-related features. Across these features, VLM-derived ratings again showed significant associations with mental health-related variables, underscoring the broad applicability of the framework to identify mental health-relevant information in real-world environments (Figure 3). For state variables of positive and negative affect, up to 33% of all context variables showed the expected direction of correlation at $p < 0.05$ (see Figure 3a), whereas for trait measures, up to 12% showed the expected pattern (Figure 3b). Together, these findings demonstrate that VLMs can accurately quantify a wide range of mental health-relevant contextual features in everyday photographs, enabling the scalability of image-based context mapping.

## Discussion

The environmental exposome is widely recognized as a fundamental determinant of human health, yet its immediate visual component has remained effectively unmeasured. While current epidemiological methods use geospatial proxies or self-report to estimate environmental exposure, these tools fail to capture the semantic content of the first-person human experience. Here, we define and quantify the visual exposome for the first time. By coupling ecological momentary assessment with VLMs, we resolve the measurement gap between macro-level environmental data and micro-level individual perception. Our results demonstrate that this framework allows for the scalable, objective decoding of associations between the visible world and affect and stress, both closely linked to mental health.

Our initial validation focused on greenness, one of the most consistently replicated environmental predictors of mental health [3]. We found that VLM-derived estimates of greenness and related features aligned robustly with participants' subjective self-reports of their visual environment and showed the expected associations with state and trait affect, as well as chronic stress. This validates that VLMs can serve as an automated proxy for human environmental assessment. Unlike satellite-derived indices, which suffer from the "uncertain geographic context problem" by capturing a location rather than the actual field of view, our approach directly evaluates what participants were visually exposed to. By analysing first-person images instead of geospatial proxies, this method provides a higher-resolution phenotype of environmental exposure that avoids the biases of self-report and the spatial coarseness of conventional metrics.

Beyond validating established associations, our findings reveal the immense complexity of the visual exposome. By developing and employing an AI-driven literature mining system to identify nearly one thousand distinct contextual features, we extended our analysis far beyond the traditional focus on greenness-associated features. The observation that a substantial proportion of these VLM-extracted context ratings significantly correlated with affective states suggests that mental health is shaped by a dense network of visual micro-exposures that have historically gone unmeasured. This represents a paradigm shift from hypothesis-driven studies of single environmental features [7] to a high-throughput, hypothesis-generating approach capable of simultaneously screening thousands of visual determinants of health. These methodological advances address a critical scalability bottleneck in psychology and public health. Analyzing the semantic content of first-person images has traditionally required labor-intensive manual coding, restricting such data to qualitative studies or small samples. By automating this process with high consistency across different model architectures, we demonstrate that visual exposomics is viable for large-scale, intensive longitudinal cohorts. This capability parallels recent technological shifts, such as wearable biosensors enabling continuous physiological monitoring at scale [22,23], and LLMs reshaping psychotherapy delivery, diabetes care [24], and personalized health coaching [25]. Just as these earlier innovations, none originally designed for their current application, reshaped how entire fields could operate, VLMs now open a comparably far-reaching frontier. This highlights the great potential of this approach to further contribute to the field of precision public health by integrating detailed contextual information into personalized and individualized models [26].

Our study has limitations. The cohort was exclusively European, limiting global generalizability, and the sample size precludes the detection of small effect sizes for rare environmental features. Furthermore, the micro-longitudinal design captures associations rather than causality. For example, positive affect may drive individuals to seek visually enriching environments as much as the environment influences affect. Future work must leverage this VLM framework in diverse populations and apply causal inference models to disentangle these bidirectional effects.

In summary, this study establishes the foundational technology for visual exposomics. We provide evidence that the human visual experience can be objectively quantified and mapped to affect and stress, both of which play central roles in mental health. This framework opens new avenues for precision public health, enabling the systematic decoding of how the visible world shapes health, from momentary states to the progression of chronic mental and physical diseases [11].

# Methods

## Study approval and participants

Ethics approval was obtained from the Ethics Committee of the University of Graz (GZ. 39/168/63 ex 2024/25). The study was conducted in accordance with institutional ethical standards. Participants were recruited via social media and personal contact. All participants provided informed consent before study enrollment and received course credits for participation.

## Ecological momentary assessment protocol

Of the 115 adults who participated, 9 participants provided data from only one alarm. The remaining 106 adults (28 men), who completed the seven-day ecological momentary assessment via the *esmira* smartphone application [21], were included in all analyses. The app delivered seven random alarms per day between 09:00 a.m. and 08:00 p.m., with a minimum interval of 30 minutes between prompts. Participants could postpone alarms if necessary. At each alarm, they reported their momentary affective well-being and the perceived greenness of their surroundings. In addition, they were instructed to capture a photograph of their current environment from their own perspective. Participants could choose to skip taking a photograph if they preferred. In total, 2,674 images with corresponding momentary self-reports were collected. Assessments began on a random weekday and continued for seven consecutive days. At the start of the ecological momentary assessment period, participants completed baseline questionnaires covering demographic and health-relevant information, including age, sex, and chronic stress.

**Mental health-related variables**

We assessed momentary affect via 10 items, which were subdivided into positive affect (5 items: active, awake, stimulated, determined, attentive) and negative affect (5 items: angry, hostile, ashamed, nervous, anxious) [27], each rated on a five-point Likert scale from 1 (not at all) to 5 (very much). Reliability analyses suggested good within-person and good between-person reliability for positive affect ($R_{Cn}$ = .64, $R_{kRn}$ = .97) and negative affect ($R_{Cn}$ = .60, $R_{kRn}$ = 97) [28]. Between-person variables (trait L level 2) were computed as each participant's mean across all momentary assessments, within-person variables (state Level 1) were the affect ratings at each alarm. The mean positive affect over all alarms was *M* = 2.89 (*SD* = 0.84) and the mean negative affect was *M* = 1.31 (*SD* = 0.49).

Additionally, we assessed momentary perceived greenness with a single item ("How green do you perceive your current environment?") rated from *not at all green* to *extremely green* on a six-point Likert scale (*M* = 2.49, *SD* = 1.51).

In addition to these momentary assessments, we collected chronic stress using the Perceived Stress Scale (PSS) [29]. The PSS measures the subjective experience of stress during the past weeks [30]. We used the validated German version [31] with ten items rated on a five-point Likert scale from 1 (never) to 5 (very often). Internal consistency was good (Cronbach's α = 0.88). The mean score was *M* = 2.90 (*SD* = 0.72).

**Vision-language model analyses for greenness and related visual features**

We primarily used the open-weight LLaMA 4 VLM (model: Maverick-17B-128E-Instruct-FP8; ≈17 billion active parameters; temperature = 0.6), accessed via a chat completion endpoint.

The model was prompted to generate quantitative ratings of greenness and visual features indicative of greenness in real-world environments, namely *greenness*, *nature score*, *plant presence,* and *natural light exposure* as well as a rating for *inside/outside* for each participant-generated photograph. To prevent potential cross-influences between categories, prompts were submitted individually, with five independent runs performed for each category. Mean scores across the five runs were used for subsequent analyses. Model outputs were parsed and stored in structured format for subsequent statistical analysis. We conducted reliability analyses similar to those performed on affect, where runs served as items. We found high within-person ($R_{Cn}$ =.97 - .99) as well as between-person ($R_{kRn}$ = .76 - .94) reliability across all runs. This argues for the calculation of mean scores for subsequent analyses. Detailed results for all five runs, including means and standard deviations, are provided in Supplementary Table 1 (full prompt texts see GitHub).

To assess the generalizability of the approach beyond a single model architecture, the same procedure, including separate prompt submission, five independent runs per category, and averaging of resulting scores, we applied Qwen3 VL (model: Qwen3-VL-235B-A22B-Instruct-FP8; ≈22 billion active parameters; temperature = 0.7). We found similar high within-person ($R_{Cn}$ = .99) as well as between-person reliability ($R_{kRn}$ = .83 - .94) scores across all runs as for the LLaMA4 VLM ratings (see Supplementary Table 1). All models were executed in November 2025 on a locally hosted high-performance computing node equipped with eight Nvidia H200 GPUs and managed through the vLLM inference engine.

**Vision-language model analyses beyond greenness**

We next examined the generalizability of the framework beyond greenness, focusing on an extensive set of contextual features that are empirically linked with mental health-related variables (positive affect, negative affect, and stress) and are potentially identifiable by VLMs in participant-generated photographs. To systematically map this conceptual space, we conducted an AI-driven, large-scale exploratory experiment that mined the entire Europe PMC database (~7 million publications) via the EPMC API using the open-weight GPT-OSS-120B model (120 billion parameters; see Supplementary Figure 2; all codes, prompts, and intermediate data files generated during the literature search and subsequent analyses are publicly available). In the first step, we conducted a keyword-based search to retrieve all publications reporting associations between the mental health-related variables of interest and contextual features potentially identifiable by VLMs. We used *Psychology* as a mandatory keyword, complemented by optional terms capturing mental health-related variables (i.e., positive affect, negative affect, stress) and contextual features (e.g., environmental, social, biological, or living elements). This search yielded 64,597 publications. In the second step, the open-weight GPT-OSS-120B model (temperature = 0.1) automatically extracted and structured relevant information from these publications, yielding 43,263 contextual features reported across 17,380 publications. In the third step, the same LLM (temperature = 0) condensed these multiword contextual features into concise one- to two-word categories suitable for VLM-based assessment. In the fourth step, after removing non-responses, seven datasets were generated, capturing all combinations of increases and decreases in positive and negative affect and stress, including one dataset containing contextual features for which studies had examined one or more mental health-related variables but reported no significant associations. In the fifth step, the LLM (temperature = 0) was used to cluster conceptually similar categories within each of these seven datasets into broader groups, resulting in 6,559 contextual features. After excluding those contextual features without significant associations, which would not allow meaningful replication within our dataset, 6,384 contextual features remained. In the sixth step, three separate datasets were created, one each for positive affect, negative affect, and stress, combining all findings on increases and decreases and removing duplicates, retaining only those contextual features for which at least three independent studies reported associations with increases or decreases in the respective mental health-related variables, yielding 1,123 contextual features, which served as the basis for the statistical analyses (for a full list of detected features for each health-relevant variable and expected effect direction see GitHub repository). After merging across the three mental health-related variables and removing duplicates, a final set of 997 unique contextual features remained.

The subsequent VLM-based ratings of the identified 997 contextual features followed the same general procedure as the greenness analyses using the open-weight LLaMA 4 VLM (model: Maverick-17B-128E-Instruct-FP8; ≈17 billion active parameters; temperature = 0.6), with the exception that only a single inference run per category was conducted for computational efficiency.

## Statistical analyses

All statistical analyses, including descriptive, correlational, and multilevel modelling procedures, were conducted in R (version 4.4.1) [32].

Using multilevel linear models implemented in the *lme4* package (version 1.1-35.5) [33], we examined the primary research questions of whether VLM-derived greenness indicators (greenness, nature score, plant presence, natural light exposure, and inside/outside) (i) corresponded to participants' self-reported greenness and whether these VLM-based measures were (ii) associated with mental health-relevant factors (i.e., affect). The VLM-derived greenness indicators (aggregated over five runs) served as dependent variables and participants included as random effects. For analysis (i), self-reported greenness was entered as the fixed predictor at Level 1 (within-person) and Level 2 (between-person). For analysis (ii), positive and negative affect were entered as fixed predictors at Level 1 (within-person) and Level 2 (between-person). In these analyses, within-person variables (state Level 1) were person-mean centered to capture fluctuations around each individual's average level, while between-person variables (trait level 2) were computed as each participant's mean across all momentary assessments. In a complementary analysis, we examined the relationship between aggregated VLM-derived greenness indicators and self-reported chronic stress, an additional mental health-related factor, by calculating Pearson product-moment correlations. To assess the generalizability of the approach beyond a single model architecture, we applied the identical analytic pipeline using the Qwen3 VL (model: Qwen3-VL-235B-A22B-Instruct-FP8 model; ≈22 billion parameters; temperature = 0.7).

To assess the scalability of the framework, we extended the analyses beyond greenness. Specifically, we examined associations between VLM-derived ratings across a broad range of contextual features, identified through the AI-based literature mining system and mental health–related psychological variables, using the same statistical approach as described above. To illustrate scalability, we computed these analyses for each single feature out of the 1,123 identified effects (described by 997 unique categories). We calculated separate analyses for positive affect (133 positive effects, 28 negative effects), negative affect (425 positive effects, 257 negative effects), and perceived stress (178 positive effects, 102 negative effects). A percentage rate of significant findings higher than 5% would argue for a detection rate of expected effects which is higher than chance.

Table 1. Multilevel model predicting objective indicators of greenness via subjective greenness ratings

| | Greenness | | | Natural light exposure | | | Plant presence | | | Nature score | | | Inside/outside | | | Average score | | |
|---|---|---|---|---|---|---|---|---|---|---|---|---|---|---|---|---|---|---|
| *Predictors* | *Estimates* | *Conf. Int (95%)* | *p-value* | *Estimates* | *Conf. Int (95%)* | *p-value* | *Estimates* | *Conf. Int (95%)* | *p-value* | *Estimates* | *Conf. Int (95%)* | *p-value* | *Estimates* | *Conf. Int (95%)* | *p-value* | *Estimates* | *Conf. Int (95%)* | *p-value* |
| Intercept | -0.04 | -0.65 – 0.57 | 0.898 | 3.58 | 2.92 – 4.25 | **<0.001** | 0.11 | -0.64 – 0.86 | 0.774 | 0.03 | -0.53 – 0.60 | 0.913 | 1.05 | 0.97 – 1.12 | **<0.001** | 0.93 | 0.34 – 1.51 | **0.002** |
| Subjective greenness trait (Level 2) | 1.28 | 1.05 – 1.51 | **<0.001** | 0.52 | 0.27 – 0.77 | **<0.001** | 1.35 | 1.07 – 1.63 | **<0.001** | 1.09 | 0.88 – 1.31 | **<0.001** | 0.08 | 0.05 – 0.11 | **<0.001** | 1.06 | 0.84 – 1.28 | **<0.001** |
| Subjective greenness state (Level 1) | 1.26 | 1.20 – 1.31 | **<0.001** | 0.75 | 0.66 – 0.83 | **<0.001** | 1.38 | 1.31 – 1.44 | **<0.001** | 1.03 | 0.97 – 1.08 | **<0.001** | 0.14 | 0.13 – 0.15 | **<0.001** | 1.10 | 1.05 – 1.15 | **<0.001** |
| **Random Effects** | | | | | | | | | | | | | | | | | | |
| $\sigma^2$ | 3.59 | | | 8.38 | | | 5.23 | | | 3.38 | | | 0.13 | | | 3.04 | | |
| $\tau_{00}$ | 0.76 $_{Participant}$ | | | 0.71 $_{Participant}$ | | | 1.17 $_{Participant}$ | | | 0.64 $_{Participant}$ | | | 0.01 $_{Participant}$ | | | 0.72 $_{Participant}$ | | |
| ICC | 0.17 | | | 0.08 | | | 0.18 | | | 0.16 | | | 0.06 | | | 0.19 | | |

| N | 106 Participant | 106 Participant | 106 Participant | 106 Participant | 106 Participant | 106 Participant |
|---|---|---|---|---|---|---|
| Observations | 2674 | 2674 | 2674 | 2674 | 2674 | 2674 |
| Marginal $R^2$ / Conditional $R^2$ | 0.453 / 0.549 | 0.108 / 0.178 | 0.400 / 0.509 | 0.381 / 0.480 | 0.223 / 0.268 | 0.419 / 0.530 |

Table 2. Multilevel model predicting VLM derived indicators of greenness via health-relevant variables

| | **Greenness** | | | **Natural light exposure** | | | **Plant presence** | | | **Nature score** | | | **Inside/outside** | | | **Average score** | | |
|---|---|---|---|---|---|---|---|---|---|---|---|---|---|---|---|---|---|---|
| *Predictors* | *Estimates* | *Conf. Int (95%)* | *p-value* | *Estimates* | *Conf. Int (95%)* | *p-value* | *Estimates* | *Conf. Int (95%)* | *p-value* | *Estimates* | *Conf. Int (95%)* | *p-value* | *Estimates* | *Conf. Int (95%)* | *p-value* | *Estimates* | *Conf. Int (95%)* | *p-value* |
| Intercept | 2.80 | 0.97 – 4.63 | **0.003** | 3.70 | 2.16 – 5.23 | **<0.001** | 2.79 | 0.67 – 4.92 | **0.010** | 2.07 | 0.46 – 3.68 | **0.012** | 1.27 | 1.09 – 1.45 | **<0.001** | 2.81 | 1.16 – 4.46 | **0.001** |
| Positive affect trait (Level 2) | 0.62 | 0.15 – 1.09 | **0.010** | 0.46 | 0.07 – 0.85 | **0.021** | 0.66 | 0.11 – 1.21 | **0.018** | 0.62 | 0.21 – 1.04 | **0.003** | 0.02 | -0.02 – 0.07 | 0.293 | 0.59 | 0.17 – 1.02 | **0.006** |
| Positive affect state (Level 1) | 0.20 | 0.05 – 0.35 | **0.008** | -0.03 | -0.21 – 0.16 | 0.786 | 0.12 | -0.06 – 0.29 | 0.189 | 0.06 | -0.08 – 0.19 | 0.422 | 0.04 | 0.02 – 0.06 | **0.001** | 0.09 | -0.05 – 0.22 | 0.206 |
| Negative affect trait (Level 2) | -1.07 | -1.88 – -0.25 | **0.010** | -0.10 | -0.78 – 0.58 | 0.773 | -0.90 | -1.85 – 0.04 | 0.061 | -0.82 | -1.54 – -0.11 | **0.024** | -0.07 | -0.15 – 0.01 | 0.095 | -0.70 | -1.43 – 0.03 | 0.061 |
| Negative affect state (Level 1) | 0.06 | -0.19 – 0.31 | 0.620 | -0.10 | -0.40 – 0.20 | 0.529 | 0.04 | -0.25 – 0.33 | 0.793 | -0.01 | -0.24 – 0.21 | 0.927 | -0.00 | -0.04 – 0.04 | 0.905 | -0.00 | -0.23 – 0.22 | 0.986 |
| **Random Effects** | | | | | | | | | | | | | | | | | | |
| $\sigma^2$ | 6.35 | | | 9.37 | | | 8.56 | | | 5.23 | | | 0.16 | | | 5.17 | | |
| $\tau_{00}$ | 1.47 Participant | | | 0.79 Participant | | | 1.99 Participant | | | 1.12 Participant | | | 0.01 Participant | | | 1.19 Participant | | |
| ICC | 0.19 | | | 0.08 | | | 0.19 | | | 0.18 | | | 0.05 | | | 0.19 | | |
| N | 106 Participant | | | 106 Participant | | | 106 Participant | | | 106 Participant | | | 106 Participant | | | 106 Participant | | |
| Observations | 2671 | | | 2671 | | | 2671 | | | 2671 | | | 2671 | | | 2671 | | |
| Marginal $R^2$ / Conditional $R^2$ | 0.033 / 0.215 | | | 0.006 / 0.084 | | | 0.022 / 0.206 | | | 0.031 / 0.202 | | | 0.008 / 0.061 | | | 0.026 / 0.209 | | |

## Figures

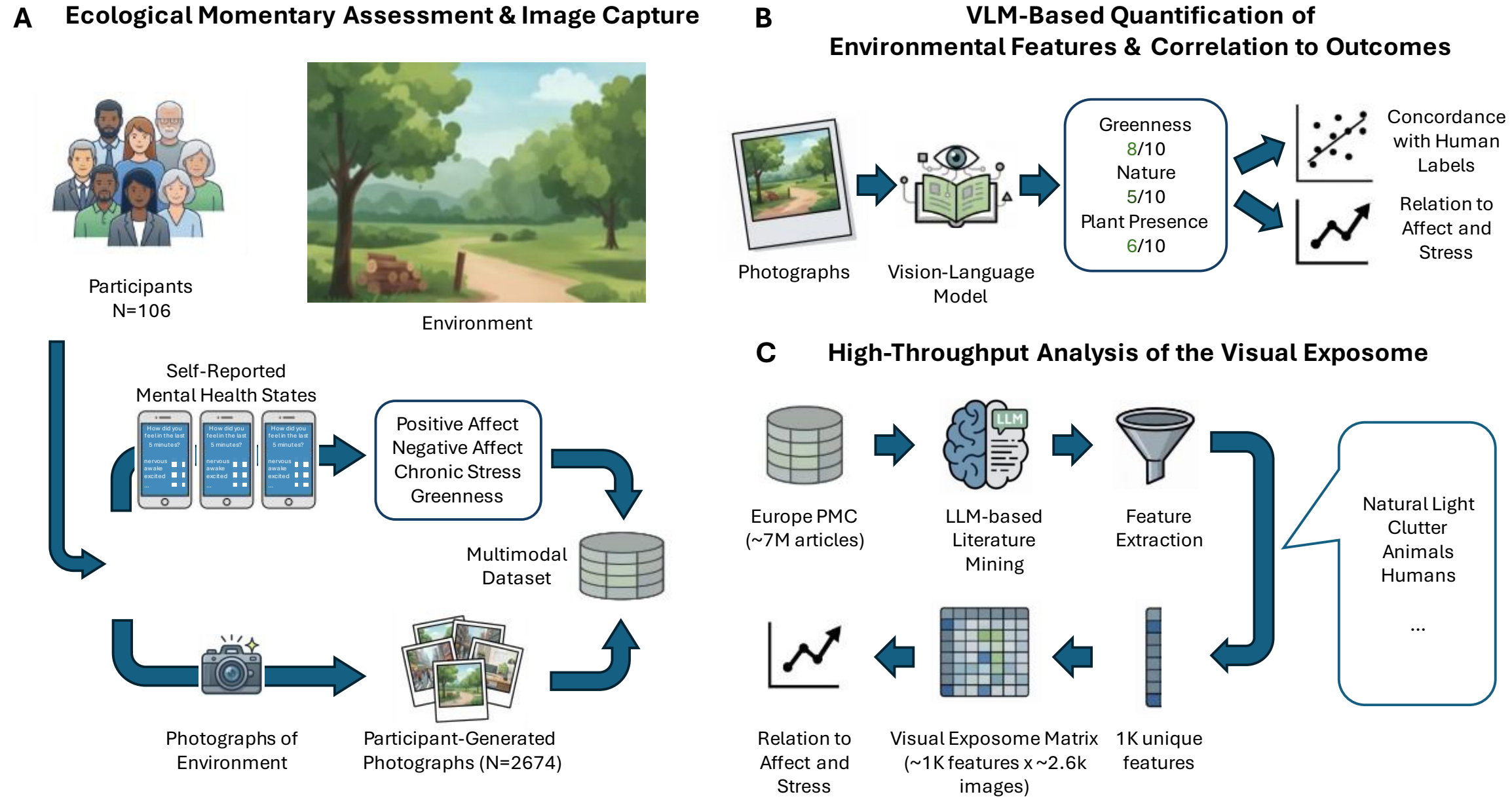

**Fig.1: Framework for first-person visual exposomics: data collection, model validation, and large-scale feature discovery. (A)** Study design and real-world data acquisition. Participants completed a 7-day ambulatory protocol, capturing 2,674 photographs that reflect their moment-to-moment visual environments alongside repeated assessments of affect and stress. **(B)** Validation of VLM-derived contextual quantification. Across 2,674 participant-generated photographs, VLM-based estimates of greenness robustly predicted momentary affect and chronic stress, replicating established associations from environmental psychology and demonstrating the validity of VLM-derived contextual measures. **(C)** Large-scale contextual feature discovery and application. A semi-autonomous LLM-driven system mined >7 million scientific publications, identifying nearly 1,000 environmental features previously linked to mental health. When applied to real-world imagery, up to 33% of VLM-extracted contextual ratings showed significant associations with affect and stress, highlighting the scalability and breadth of VLM-based visual exposomics.

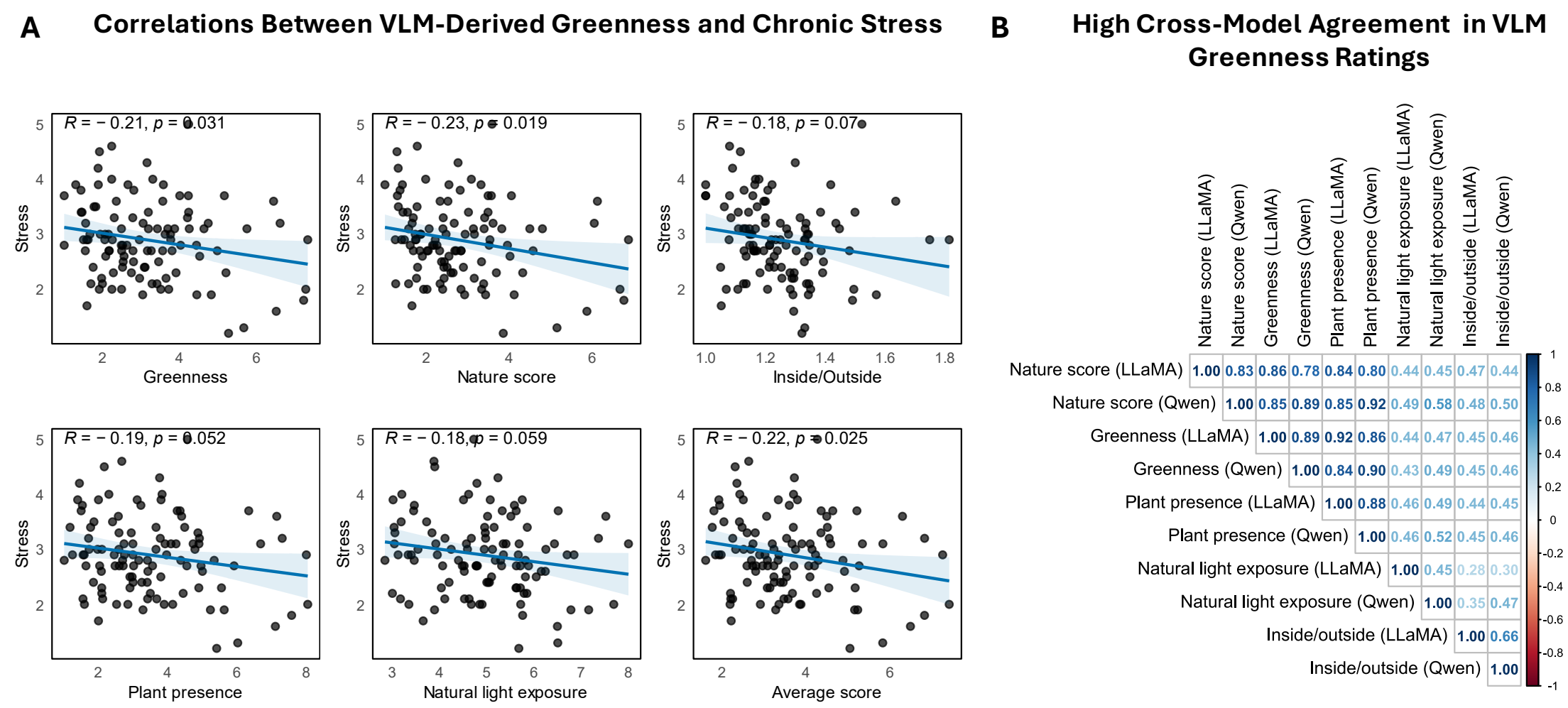

**Fig.2: Greenness-stress associations and cross-VLM consistency in contextual ratings. (A)** Trait-level associations between vision language model (VLM)-derived greenness features and chronic stress. Higher greenness was significantly associated with lower chronic stress, with all greenness features showing negative correlations of comparable magnitude. **(B)** Consistency of greenness-related feature ratings across two VLMs (LLaMA 4 VLM and open-weight Qwen3 VL). Both models produced highly similar ratings for greenness, nature score, and plant presence, with especially high consistency for nature score and greenness.

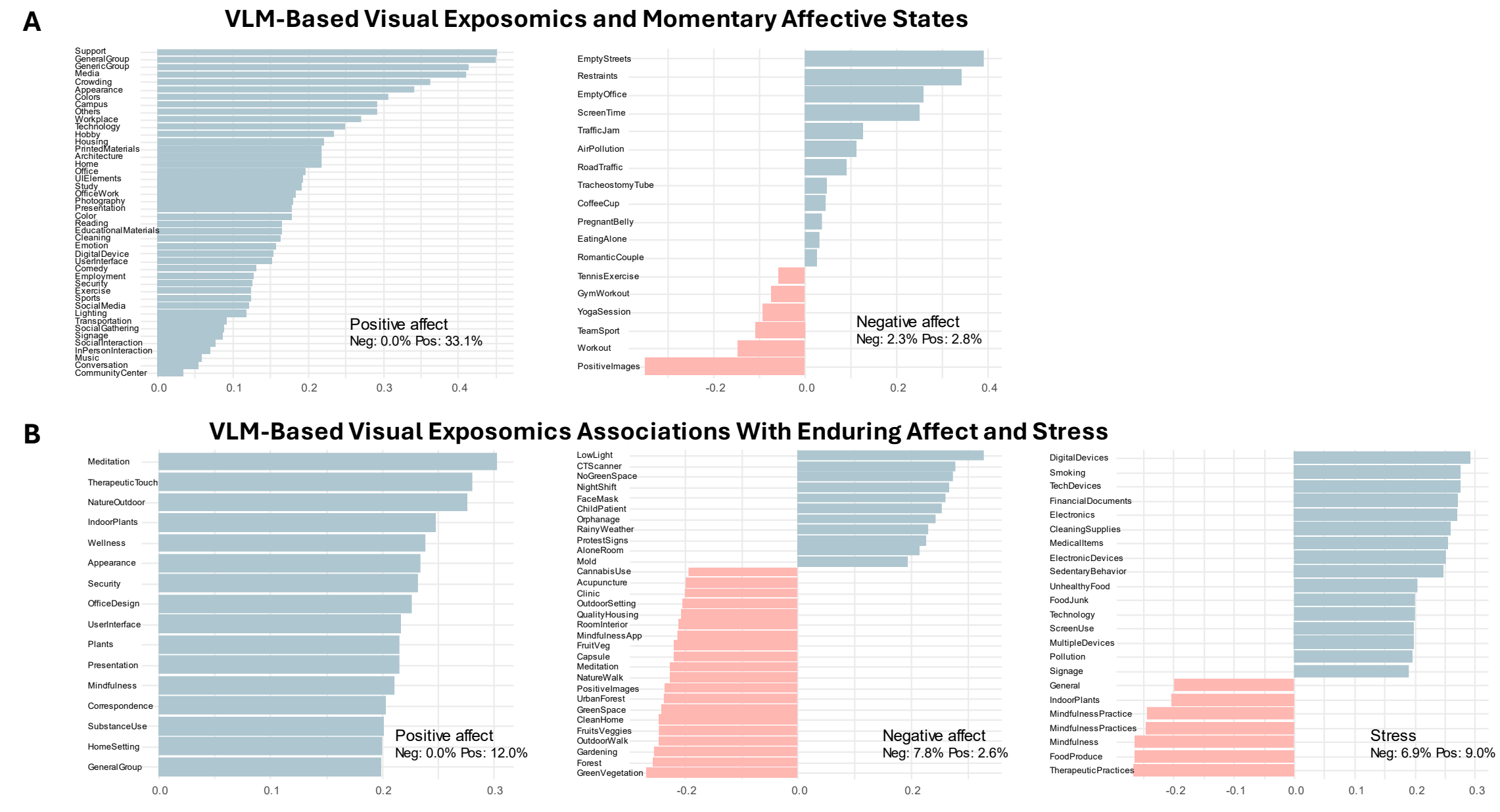

**Fig.3: Associations between the literature-derived visual exposome, affect and stress.** Significant associations were defined as $p < 0.05$. Positive associations are shown in blue, negative associations in pink. **(A)** Literature-derived visual exposome features associated with momentary affect and stress. **(B)** Literature-derived visual exposome features associated with enduring affect and stress.

**Funding**

J.N.K. is supported by the German Cancer Aid DKH (DECADE, 70115166), the German Federal Ministry of Research, Technology and Space BMFTR (PEARL, 01KD2104C; CAMINO, 01EO2101; TRANSFORM LIVER, 031L0312A; TANGERINE, 01KT2302 through ERA-NET Transcan; Come2Data, 16DKZ2044A; DEEP-HCC, 031L0315A; DECIPHER-M, 01KD2420A; NextBIG, 01ZU2402A), the German Research Foundation DFG (TRR 412/1, 535081457; SFB 1709/1 2025, 533056198), the German Academic Exchange Service DAAD (SECAI, 57616814), the German Federal Joint Committee G-BA (TransplantKI, 01VSF21048), the European Union EU's Horizon Europe research and innovation programme (ODELIA, 101057091; GENIAL, 101096312), the European Research Council ERC (NADIR, 101114631), the Breast cancer Research Foundation (BELLADONNA, BCRF-25-225) and the National Institute for Health and Care Research NIHR (Leeds Biomedical Research Centre, NIHR203331). The views expressed are those of the author(s) and not necessarily those of the NHS, the NIHR or the Department of Health and Social Care. This work was funded by the European Union. Views and opinions expressed are, however, those of the author(s) only and do not necessarily reflect those of the European Union. Neither the European Union nor the granting authority can be held responsible for them. No additional funding was received for this study beyond the support listed above.

**Role of the funding source**

The funders had no role in the study design, data collection, data analysis, data interpretation, or writing of the report.

## Data availability

Correspondence and requests should be addressed to the corresponding authors. Raw data generated by the LLM-based literature search are available at the project's GitHub repository (https://github.com/WekenborgLab/AmbuVision.git).

Outputs from the VLM analyses that were used for statistical modelling, together with anonymized participant-level data, are available on OSF (https://osf.io/jm5ba/overview?view_only=3a946cc1b16a49efa7d41a9e99604657).

Original participant photographs cannot be shared due to privacy protections, as identifiable individuals and sensitive contextual information may be visible in the images.

## Code availability

All analyses were performed using publicly available software. All source codes as well as complete R code for replication of all data analyses and figures are available at https://osf.io/jm5ba/overview?view_only=3a946cc1b16a49efa7d41a9e99604657.

## Declaration of interest

JNK declares ongoing consulting services for AstraZeneca and Bioptimus. Furthermore, he holds shares in StratifAI, Synagen, and Spira Labs, has received an institutional research grant from GSK and AstraZeneca, as well as honoraria from AstraZeneca, Bayer, Daiichi Sankyo, Eisai, Janssen, Merck, MSD, BMS, Roche, Pfizer, and Fresenius.

## Author Contributions

C.R., M.K.W., and J.N.K. conceived the study. C.R. collected the human behavioral data, performed the statistical analyses, and drafted the manuscript. A.R.S. contributed to human data collection and supervised the experimental procedures. M.G.S., D.K., E.A.M.M., and F.W. designed and implemented the prompt engineering for the large language model (LLM)- based vision-language analyses and curated the LLM-generated output. J.N.K. conceived the overall project framework, supervised the LLM analysis pipeline, and contributed to manuscript conceptualization. M.K.W. contributed to manuscript conceptualization, prompt engineering, and supervised the overall project. All authors critically revised the manuscript for important intellectual content and approved the final version.

**Inventory of Supplemental Information**

*Supplementary Table 1.* Original and mean values of VLM-greenness-related image metrics over 5 runs for LLaMA4 VLM and Qwen3 VL

*Supplementary Table 2.* Multilevel model predicting VLM-derived greenness indicators from the Qwen3 VL model via health-related factors

*Supplementary Figure 1.* Associations between chronic stress and greenness indicators derived from Qwen3 VL

*Supplementary Figure 2.* AI-driven workflow of the literature-derived visual exposome

Supplementary Table 1. Original and mean values of VLM-greenness-related image metrics over 5 runs for LLaMA4 VLM and Qwen3 VL

| | **Greenness** | | | **Natural light exposure** | | | **Plant presence** | | | **Nature score** | | | **Inside/outside** | | | |
|---|---|---|---|---|---|---|---|---|---|---|---|---|---|---|---|---|
| | M | SD | Min-max | M | SD | Min-max | M | SD | Min-max | M | SD | Min-max | M | SD | Min-max | |
| Run 1 | 3.15 | 2.86 | 1-10 | 4.88 | 3.36 | 1-10 | 3.45 | 3.29 | 1-10 | 2.77 | 2.61 | 1-9 | 1.25 | 0.43 | 1-2 | Llama 4 VLM |
| Run 2 | 3.13 | 2.84 | 1-10 | 4.85 | 3.36 | 1-10 | 3.45 | 3.30 | 1-10 | 2.75 | 2.59 | 1-9 | 1.25 | 0.43 | 1-2 | |
| Run 3 | 3.14 | 2.85 | 1-10 | 4.88 | 3.37 | 1-10 | 3.45 | 3.30 | 1-10 | 2.75 | 2.59 | 1-9 | 1.25 | 0.43 | 1-2 | |
| Run 4 | 3.13 | 2.85 | 1-10 | 4.86 | 3.36 | 1-10 | 3.45 | 3.29 | 1-10 | 2.74 | 2.59 | 1-9 | 1.24 | 0.43 | 1-2 | |
| Run 5 | 3.13 | 2.86 | 1-10 | 4.84 | 3.37 | 1-10 | 3.44 | 3.30 | 1-10 | 2.75 | 2.60 | 1-10 | 1.25 | 0.43 | 1-2 | |
| Mean | 3.13 | 2.83 | 1-10 | 4.86 | 3.19 | 1-10 | 3.45 | 3.26 | 1-10 | 2.75 | 2.54 | 1-9 | 1.25 | 0.41 | 1-2 | |
| Mean confidence | 9.13 | 1.04 | 2.2-10 | 8.84 | 1.02 | 2.8-10 | 9.44 | 0.85 | 2.8-10 | 9.02 | 0.90 | 6-10 | 8.15 | 0.54 | 1-10 | |
| Run 1 | 3.61 | 2.93 | 1-10 | 6.16 | 2.29 | 1-10 | 3.76 | 3.40 | 1-10 | 3.23 | 2.64 | 1-10 | 1.20 | 0.40 | 1-2 | Qwen3 VL |
| Run 2 | 3.61 | 2.93 | 1-10 | 6.17 | 2.31 | 1-10 | 3.76 | 3.41 | 1-10 | 3.24 | 2.64 | 1-10 | 1.20 | 0.40 | 1-2 | |
| Run 3 | 3.61 | 2.93 | 1-10 | 6.18 | 2.28 | 1-10 | 3.78 | 3.40 | 1-10 | 3.23 | 2.64 | 1-10 | 1.20 | 0.40 | 1-2 | |
| Run 4 | 3.61 | 2.93 | 1-10 | 6.18 | 2.29 | 1-10 | 3.76 | 3.40 | 1-10 | 3.23 | 2.64 | 1-10 | 1.20 | 0.40 | 1-2 | |
| Run 5 | 3.61 | 2.93 | 1-10 | 6.17 | 2.31 | 1-10 | 3.76 | 3.39 | 1-10 | 3.23 | 2.64 | 1-10 | 1.20 | 0.40 | 1-2 | |
| Mean | 3.61 | 2.92 | 1-10 | 6.17 | 2.27 | 1-10 | 3.76 | 3.39 | 1-10 | 3.23 | 2.63 | 1-10 | 1.20 | 0.40 | 1-2 | |
| Mean confidence | 9.13 | 0.95 | 3-10 | 8.47 | 0.82 | 4-10 | 9.47 | 0.77 | 3.6-10 | 9.01 | 0.92 | 4-10 | 9.79 | 0.43 | 5-10 | |

Supplementary Table 2. Multilevel model predicting VLM-derived greenness indicators from the Qwen3 VL model via health-related factors.

| | **Greenness** | | | **Natural light exposure** | | | **Plant presence** | | | **Nature score** | | | **Inside/outside** | | | **Average score** | | |
|---|---|---|---|---|---|---|---|---|---|---|---|---|---|---|---|---|---|---|
| *Predictors* | *Estimates* | *Conf. Int (95%)* | *p-value* | *Estimates* | *Conf. Int (95%)* | *p-value* | *Estimates* | *Conf. Int (95%)* | *p-value* | *Estimates* | *Conf. Int (95%)* | *p-value* | *Estimates* | *Conf. Int (95%)* | *p-value* | *Estimates* | *Conf. Int (95%)* | *p-value* |
| Intercept | 3.30 | 1.37 – 5.24 | **0.001** | 5.24 | 3.97 – 6.52 | **<0.001** | 3.18 | 0.87 – 5.50 | **0.007** | 2.54 | 0.75 – 4.33 | **0.005** | 1.23 | 1.03 – 1.43 | **<0.001** | 3.57 | 1.89 – 5.25 | **<0.001** |
| Positive affect trait (Level 2) | 0.60 | 0.10 – 1.10 | **0.018** | 0.44 | 0.11 – 0.77 | **0.008** | 0.64 | 0.04 – 1.24 | **0.035** | 0.65 | 0.19 – 1.12 | **0.006** | 0.02 | -0.03 – 0.07 | 0.412 | 0.59 | 0.15 – 1.02 | **0.008** |
| Positive affect state (Level 1) | 0.10 | -0.05 – 0.25 | 0.193 | 0.19 | 0.07 – 0.32 | **0.002** | -0.01 | -0.18 – 0.17 | 0.945 | 0.05 | -0.08 – 0.19 | 0.434 | 0.07 | 0.04 – 0.09 | **<0.001** | 0.09 | -0.05 – 0.22 | 0.201 |
| Negative affect trait (Level 2) | -1.06 | -1.92 – -0.20 | **0.016** | -0.23 | -0.79 – 0.33 | 0.424 | -0.91 | -1.94 – 0.12 | 0.082 | -0.87 | -1.66 – -0.07 | **0.032** | -0.07 | -0.15 – 0.02 | 0.141 | -0.77 | -1.51 – -0.02 | **0.044** |
| Negative affect state (Level 1) | 0.06 | -0.19 – 0.31 | 0.650 | -0.04 | -0.24 – 0.17 | 0.725 | 0.09 | -0.21 – 0.38 | 0.559 | 0.02 | -0.20 – 0.25 | 0.841 | -0.02 | -0.06 – 0.01 | 0.208 | 0.03 | -0.19 – 0.25 | 0.767 |
| **Random Effects** | | | | | | | | | | | | | | | | | | |
| $\sigma^2$ | 6.64 | | | 4.43 | | | 8.97 | | | 5.30 | | | 0.14 | | | 4.94 | | |
| $\tau_{00}$ | 1.68 $_{Participant}$ | | | 0.64 $_{Participant}$ | | | 2.43 $_{Participant}$ | | | 1.46 $_{Participant}$ | | | 0.01 $_{Participant}$ | | | 1.27 $_{Participant}$ | | |
| ICC | 0.20 | | | 0.13 | | | 0.21 | | | 0.22 | | | 0.08 | | | 0.20 | | |
| N | 106 $_{Participant}$ | | | 106 $_{Participant}$ | | | 106 $_{Participant}$ | | | 106 $_{Participant}$ | | | 106 $_{Participant}$ | | | 106 $_{Participant}$ | | |
| Observations | 2671 | | | 2671 | | | 2671 | | | 2671 | | | 2671 | | | 2671 | | |
| Marginal $R^2$ / Conditional $R^2$ | 0.029 / 0.225 | | | 0.016 / 0.140 | | | 0.020 / 0.229 | | | 0.032 / 0.241 | | | 0.015 / 0.098 | | | 0.028 / 0.226 | | |

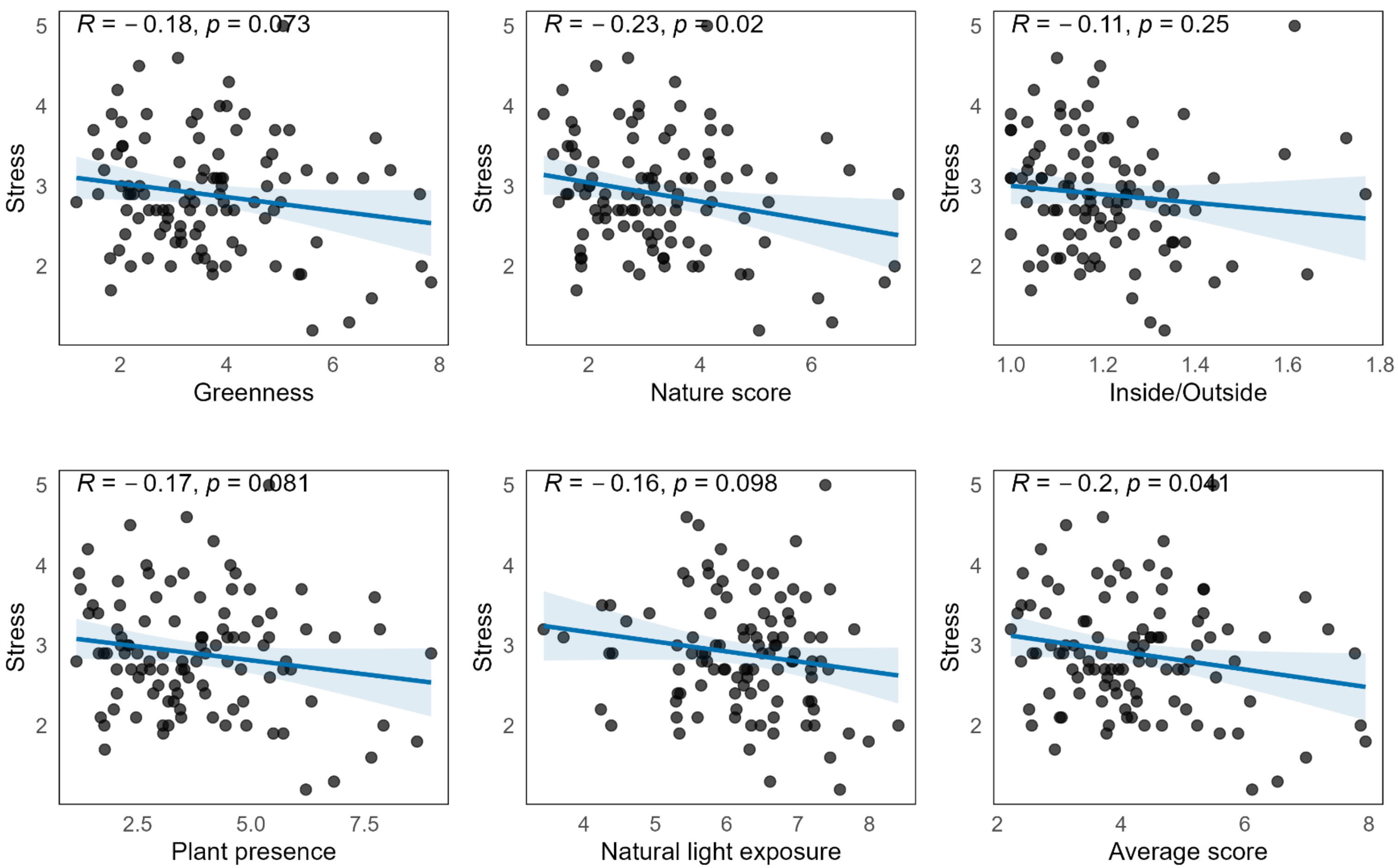

**Supplementary Fig.1: Greenness-stress associations based on greenness estimates derived from Qwen3 VL.** Trait-level associations between **Qwen3 VL-derived greenness estimates** and chronic stress. Higher greenness was associated with lower chronic stress, with all greenness indicators showing negative correlations of comparable magnitude

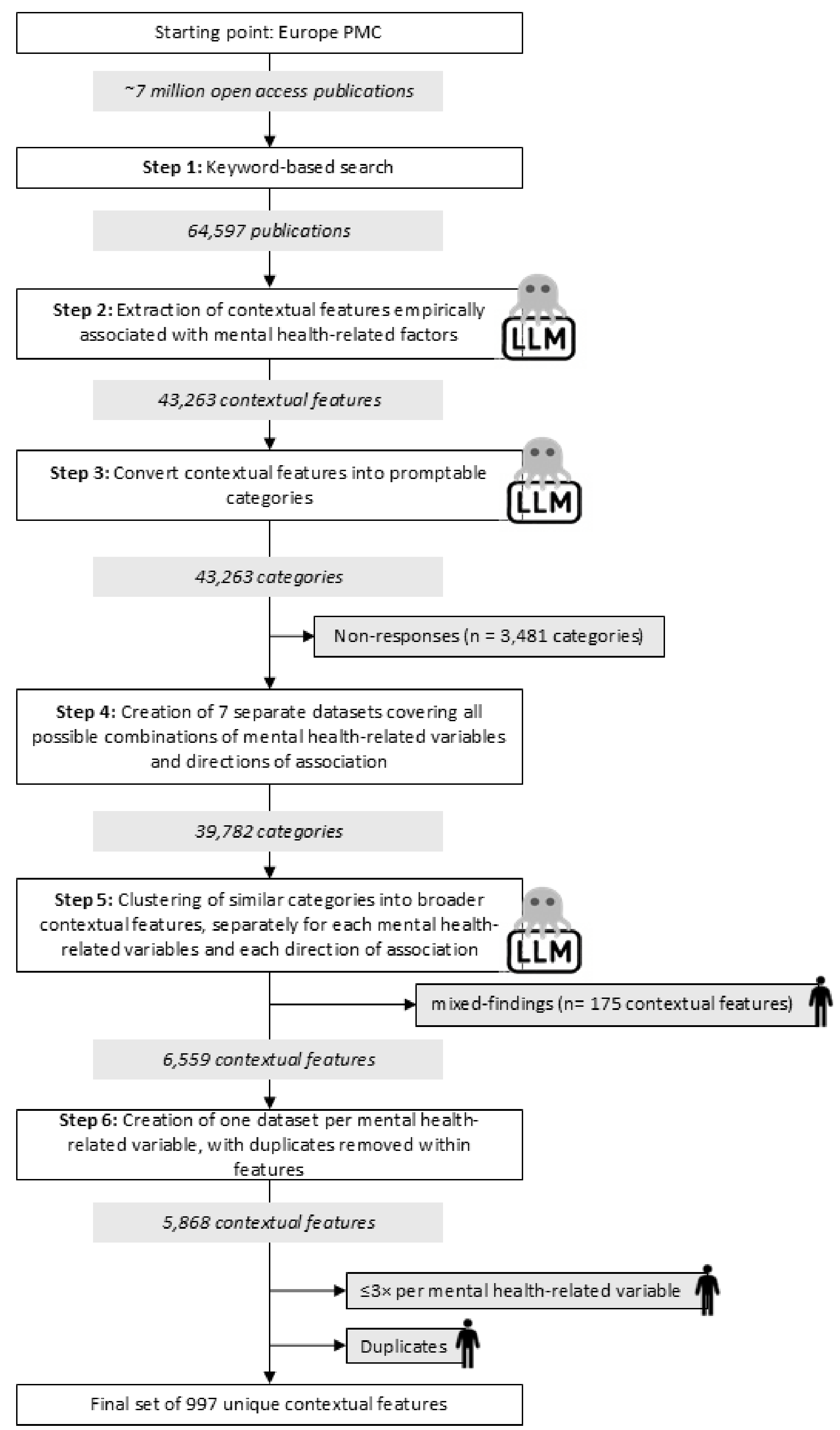

**Supplementary Fig.2: AI-Driven Workflow of the Literature-Derived Visual Exposome**. The figure illustrates the AI-driven workflow used to extract and organize environmental contextual features from the scientific literature. A large language model (LLM) processes millions of full-text publications to identify variables relevant to mental health and group them into coherent categories that collectively define the literature-derived visual exposome used in the present study. Pictograms of a person indicate human-in-the-loop steps involving targeted manual oversight or refinement.